\pdfoutput=1
\documentclass[11pt]{article}

\usepackage[preprint]{acl}

\usepackage{times}
\usepackage{latexsym}
\usepackage{amsmath} 
\usepackage{subfigure}
\usepackage{booktabs}
\usepackage{hyperref}
\usepackage{authblk}
\usepackage{listings}
\usepackage{float}
\usepackage[T1]{fontenc}
\usepackage{microtype}
\usepackage{inconsolata}
\usepackage{graphicx}

\title{Topic Segmentation Using Generative Language Models}

\author{
 \textbf{Pierre Mackenzie}$^{\star,\dagger}$,
 \textbf{Maya Shah}$^{\star,\ddagger}$,
 \textbf{Patrick Frenett}$^{\star}$,
\\
 $^\star$Adarga,
 $^\dagger$University of Edinburgh,
 $^\ddagger$University of Exeter,
\\
 \small{
   \textbf{Correspondence:} \texttt{lardet[dot]pierre[at]gmail.com}
 }
}

\begin{document}
\maketitle
\begin{abstract}
  Topic segmentation using generative Large Language Models (LLMs) remains relatively unexplored. Previous methods use semantic similarity between sentences, but such models lack the long range dependencies and vast knowledge found in LLMs. In this work, we propose an overlapping and recursive prompting strategy using sentence enumeration. We also support the adoption of the boundary similarity evaluation metric. Results show that LLMs can be more effective segmenters than existing methods, but issues remain to be solved before they can be relied upon for topic segmentation.
\end{abstract}

\section{Introduction}

% 1
% ---------- Introduction ----------
% 1.1 Motivation
% \subsection{Motivation}

% Topic segmentation can be important as a pre-processing step before some other NLP task, or can be important in its own right.

Topic segmentation is the problem of dividing a string of text into constituent `segments'. Each segment should be semantically self-contained such that it is about one thing. For this work, segment boundaries always lie on sentence boundaries. We can then interpret segmentation as a binary classification task; given a list of input sentences, the model must decide whether there exists a boundary between each pair of adjacent sentences.

Despite advances in LLMs, topic segmentation remains a relevant task. Information retrieval, long-document summarisation, classification and RAG~\cite{RAG} can all benefit from their inputs first being broken down by topic. For many open-source or resource-constrained models, context windows limit the size of input. Although newer models have very long context windows, \citet{EffectOfLongContextWindows} show that LLMs do not fully utilise long context. Segmentation can also be important for its own sake in dividing a document into constituent parts, to create a contents page, or summaries and titles for each section.

Segmentation is a non-trivial task. The ambiguous definition of a segment leads to disagreements between humans annotators on where the `correct' boundaries lie~\citep{TextTiling}. This is perhaps why there are few datasets in the field and none with human annotations of passages of text. Instead, annotations are normally derived from concatenations or metadata. For a machine learning model to attain performance comparable to humans is a daunting task that is hard to measure.

\begin{figure*}[t]
    \includegraphics[width=\textwidth]{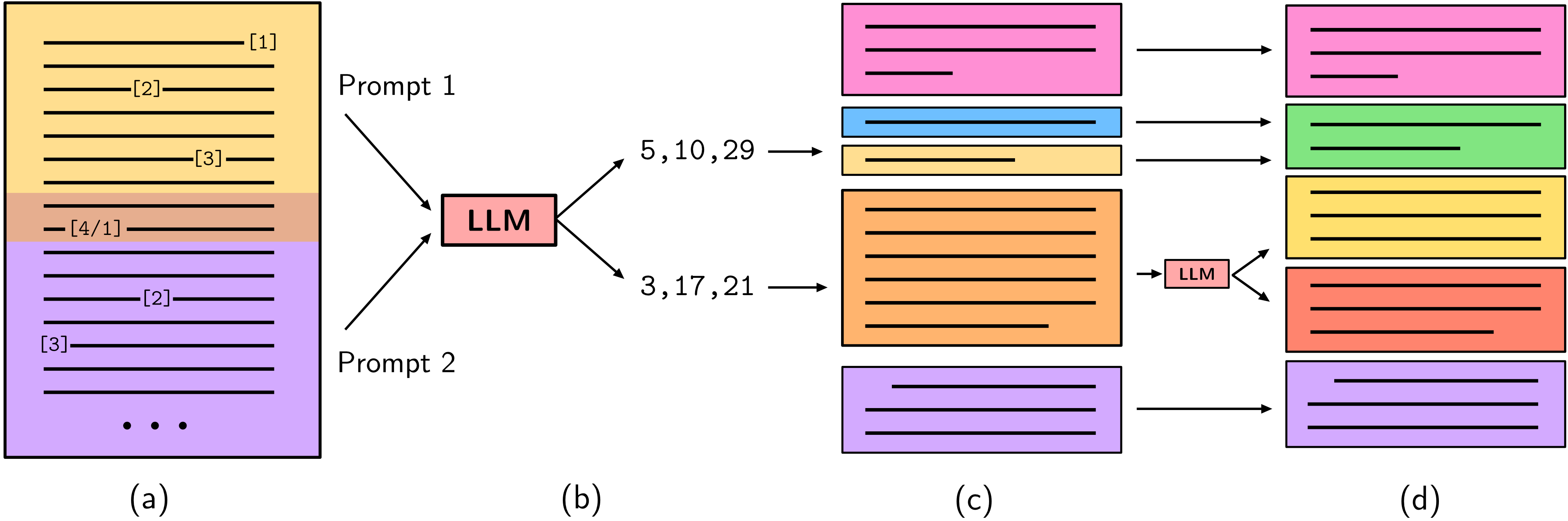}
    \caption{The overlapping and recursive prompting strategy for segmentation. In (a), a long document is split into overlapping sections with sentence boundaries enumerated. In (b), each section is segmented by the LLM. In (c), the segments are joined. Finally, in (d), each segment is validated to ensure it is not too long or too short.}~\label{fig:diagram}
    \vspace{-2em}
  \end{figure*}
\section{Related Work}

% Please refer to \citep{XingThesis} which provides a recent, broad overview of topic segmentation.

An influential framework introduced by~\citep{TextTiling} involves computing lexical similarity scores between adjacent sentences before boundaries are placed where similarity is lowest~\citep{lexical1,lexical2}. Such a framework is still in use today but with semantic similarity calculated from embeddings. Different neural architectures have seen use such as RNNs ~\citep{BiLSTM,HierarchicalBiLSTM,CNNFeaturesLSTMAttention} and attention-based models~\cite{CrossAttentionHierarchical,TwoLevelTransformerSoftmax,TwoLevelTransformerPretrained}. However, these models do not leverage the vast knowledge contained in the largest pre-trained language models.

LLMs are the state of the art for a variety of NLP tasks. However, there has been little research into their use for topic segmentation. A loss-based approach was proposed by~\citep{DialoGPT} in which boundaries are placed at peaks in the mean negative log likelihood of tokens in a sentence. This method relies on the dubious assumption that all information about segment boundary location can be expressed by the next token prediction loss. Due to resource constraints, we would have to use less powerful LLMs to test this method, and \citet{XingThesis} finds that prompting LLMs outperforms this method. Therefore, we do not consider it here.

Previous unpublished work has explored segmentation by prompting LLMs~\citep{XingThesis}. They find that prompting ChatGPT is the best dialogue segmentation model unless the input exceeds ChatGPT's limit. Two prompting methods are proposed: one which asks the LLM to return the original text with characters delimiting boundaries and a second in which the LLM is asked to return semantic coherence scores for each pair of sentences in the range $(0,1)$. The first method does not satisfy a guarantee that the model will return the original document unedited and is massively wasteful of tokens. In the second method, the returned scores have no guarantee of directly corresponding to semantic coherence specifically for topic segmentation. Due to monetary constraints, we were unable to test more than one prompting method, and therefore we do not empirically compare against this work.

We propose a new prompting method which ensures the output text is unedited, is vastly more token-efficient, and is not limited by the context window of the model. We show that prompting method outperforms existing non-prompting approaches by comparison with semantic similarity calculated on SentenceBERT~\citep{SentenceBERT} embeddings.

% \vspace{-0.2em}
\begin{table*}[t]
    \centering
    \begin{tabular}{@{}lcccc@{\hspace{2em}}ccccc@{\hspace{0.2em}}c@{}}
    \toprule
    & \multicolumn{4}{c}{boundary similarity ($n=2$)} & \multicolumn{6}{c}{boundary precision and recall} \\ 
    \midrule
    & Human & Wiki & Conc-Wiki & Synthetic & \multicolumn{2}{c}{Human} & \multicolumn{2}{c}{Wiki} & \multicolumn{2}{c}{Conc-Wiki} \\ 
    &  &  &  &  & BP & BR & BP & BR & BP & BR \\
    \midrule
    \it GPT3.5  & \bf 0.38 & \bf 0.25 & 0.29 & \bf 0.35 & \bf 0.51 & \bf 0.60 & 0.36 & \bf 0.55 & 0.42 & \bf 0.63 \\ 
    \it FlanT5  & 0.25 & 0.24 & 0.41 & 0.33 & 0.38 & 0.46 & \bf 0.43 & 0.37 & 0.65 & \bf 0.63 \\ 
    \midrule
    \it BERTGraph   & 0.20 & 0.15 & 0.45 & 0.21 & 0.47 & 0.25 & 0.39 & 0.21 & 0.79 & 0.54 \\ 
    \it BERT    & 0.18 & 0.09 & \bf 0.46 & 0.18 & 0.33 & 0.39 & 0.23 & 0.37 & \bf 0.91 & 0.50 \\ 
    \it Split5  & 0.13 & 0.19 & 0.19 & 0.23 & 0.17 & 0.48 & 0.33 & 0.39 & 0.25 & 0.52 \\ 
    \bottomrule
    \end{tabular}
    \caption{Mean boundary similarity, precision (BP) and recall (BR) with $n=2$ for each model computed on a maximum of 150 documents per dataset. Note that \emph{Human} has only 10 documents and \emph{Synthetic} annotations were generated by ChatGPT. Best results for each metric/dataset are in bold. The LLMs perform better on all datasets except \emph{Conc-Wiki}. They have better recall whereas precision varies per dataset.}~\label{tab:combined_results}
    \vspace{-2em}
\end{table*}

% 2
% ---------- Method ----------
\section{Method}
% 2. Method
% 2.1 Datasets
\subsection{Datasets}

% There are 4 datasets used in the experiments in this work: a small human-annotated dataset, a scrape of English wikipedia, a `concatenated' wikipedia scrape and a synthetic GPT3.5 generated dataset.

\hspace*{\parindent}\emph{Human}: We use a small dataset of 10 documents manually segmented by the authors. It is comprised of miscellaneous non-fiction articles. This was intended to provide high quality examples for qualitative analysis.

\emph{Wiki}: A wikipedia scrape\footnote{\url{https://www.kaggle.com/datasets/ltcmdrdata/plain-text-wikipedia-202011/data}} was automatically segmented based on headings and filtered to remove articles with too few segments (<4), too short segments (<20 words) or too many artefacts (> 20\% non-alphabetic characters), leaving \textasciitilde{}1000 articles. We evaluate without headings present.

\emph{Conc-Wiki}: We randomly sampled segments from \emph{Wiki} and concatenated them to form new incoherent articles, with segments drawn from completely different domains, leading to \textasciitilde{}500 articles.

\emph{Synthetic}: Segmentations were generated by \emph{GPT-3.5}~(\ref{GPT3.5}) on proprietary source data consisting of technical/news reports. This dataset was used for both fine-tuning \emph{FlanT5} (\ref{FlanT5}) and evaluation.
% 2. Method
% 2.2 Evaluation
\subsection{Evaluation}\label{evaluation}

We follow the work of \citet{fournier-2013-B} who propose the `boundary similarity' metric and associated precision/recall. The metric pairs segment boundaries between a hypothesised and references segmentation. Exact matches score 1 and no match scores 0, whilst matches within a distance $n$ score linearly in the distance. Boundary similarity (B) is the mean score, while boundary precision/recall (BP/BR) are the mean score of matched hypothesis/reference boundaries, respectively. For further justification for the use of boundary similarity as opposed to more traditional metrics such as WD and Pk~\cite{HearstW2002}, see the work of \citet{fournier-2013-B}, or our own investigations\footnote{redacted to retain anonymity}.
% \url{https://github.com/PierreRL/segmenter-evaluation-metrics}.
% 2.3 LLM-Based Text Segmentation
\subsection{LLM-Based Text Segmentation}\label{LLM-Based Text Segmentation}

How can we get LLMs to output segment boundaries? We might consider prompting with the input text and asking the LLM to add characters delimiting boundaries~\citep{XingThesis}. However, not only is this wasteful of tokens, but the LLM may fail to copy the input perfectly. These problems are addressed by~\citep{XingThesis} through repeated prompting until sequence lengths match, but this provides no guarantees. Our use case requires a guarantee that the input data would be unedited, so we opt for a different prompting strategy. The method is illustrated in Figure~\ref{fig:diagram} and is described below.

We first annotate the text with indices between each sentence. As an example: `Hello World. [1] The sky is blue. [2] The sun is is yellow'. We then ask the LLM to return a list of indices corresponding to boundaries. In the previous example, the ideal response might be `1'. We add a system prompt which describes the segmentation task, desired output format and primes the model for segmentation, exemplified in Appendix~\ref{Prompting Strategy}. We also add a variety of examples in line with the few-shot prompting technique~\citep{FewShotLearners}. 

% This did not necessarily reflect the fact that the LLM learned how to segment better. Instead, through manual testing, we suspect that it learned the ideal segment length and amount of information that should be contained within a segment, which was implicitly contained in the few-shot prompt (and also the testing datasets, therefore increasing performance). This suggests that different implicit definitions of a segment could be imposed by a few-shot prompt to an LLM.

Many of our input documents exceeded the context window of models available at the time. Therefore, we propose a simple overlapping prompt strategy to overcome this limitation. This should not be done by splitting the text at the sentence nearest to the context window limit for two reasons. First, we do not know whether this sentence boundary should serve as a segment boundary. Second, the LLM loses valuable context which helps to choose where to place boundaries at the extremes. Therefore, we send prompts with some overlap between sections. We choose an overlap of twice the maximum segment length. In our experiments, we set a maximum segment length of 750 tokens and hence an overlap of 1500 tokens. Given two generations which were prompted by 1500 overlapping tokens, we accept boundaries for the first 750 tokens from the first prompt and the boundaries in the final 750 tokens from the second. 

% Our prompting method works so long as the input text is within the context window of the LLM. At the time of experimentation, and with the use of gpt-3.5-turbo, we had a limit of 16k tokens. Many of our input texts exceeded this limit. Therefore, we need to split up these long documents into smaller chunks that can be processed by the LLM.

% We did not experiment with involving the responses from both outputs, but found no sign of worse performance towards segment boundaries.

% While this method is wasteful of up to 1500 tokens per prompt, this is a small enough fraction of the 16k context that we were satisfied with the solution. If maximum segment lengths were much longer, the context may need to be limited more severely.

We also perform validation on the segments returned by the LLM. We first verify that the returned segments are of an appropriate length. Segments that are too short are concatenated with a neighbouring segment and segments that are too long are recursively segmented by the same model. Recursive segmentation was done with another prompt that asks the model to generate a single boundary. Again, we use a few-shot prompting strategy. We choose a minimum segment length of 50 words and a maximum of 500 words.

% 3. Experiments
\section{Experiments}\label{Results}
% 3.1 Results
\subsection{Models}
\hspace*{\parindent}\emph{Split5}: A simple baseline which creates segment boundaries every 5 sentences.

\emph{BERT}: Generates a sequence of similarities using embeddings from SentenceBERT. Each element is a weighted average of the cosine similarities with the previous $5$ sentences. Boundaries are placed at troughs in the sequence before long segments are split and short segments are grouped. The threshold for boundary placement is set to 0.3 based on manual testing.

\emph{BERTGraph}: \citet{MasimilianoSegmenter} also uses SentenceBERT cosine similarities, but cluster sentences as a graph to find segments. Post-processing ensures that segments are contiguous.

\emph{GPT3.5}\label{GPT3.5}: OpenAI's \texttt{gpt-3.5-turbo-16k}\footnote{Queries were made in August 2023.} was queried using the method defined in Section~\ref{LLM-Based Text Segmentation}. We choose to use deterministic outputs from the model by setting topk $=1$. 

\emph{FlanT5}\label{FlanT5}: We fine-tune Google's Flan-T5 large~\citep{FlanT5} (780M) using LORA~\citep{LORA} on a combination of \emph{Wiki}, \emph{Conc-Wiki} and a training split of \emph{Synthetic} which took \textasciitilde{}24 GPU hours. We use the same topk $=1$ as \emph{GPT3.5}. Because the model was fine-tuned, we use a much shorter prompt and no few-shot examples.
% 3.2 Results

\subsection{Quantitative Results}

Models were tested on \emph{Human}, \emph{Wiki}, \emph{Conc-Wiki} and a test-partition of \emph{Synthetic}. We do not base our conclusions on results from \emph{Synthetic} as they would be biased in favor of the generative models which generated the segmentations. We evaluate using the previously discussed boundary similarity~\ref{evaluation} with $n=2$, as the metric becomes noisier with higher values of $n$. Results can be found in Table~\ref{tab:combined_results}.

Due to resource constraints, we could not test \emph{GPT3.5} on the full \emph{Wiki} or \emph{Conc-Wiki} datasets. Instead, we report results from the largest subset that fit within resource constraints. This was 150 documents per dataset. We evaluated all other models on the full datasets to verify that relative performance across models is similar. Full results on all evaluation metrics can be found at the following links for both the \href{https://docs.google.com/spreadsheets/d/e/2PACX-1vT3XnZ-npYMquwabYd_WGZrvFLtNDTsN-qwp94-kKR4M6Fq0Y0f87a6P0RNQ_W1VvGJtE_kPI5conlA/pubhtml}{subset} and \href{https://docs.google.com/spreadsheets/d/e/2PACX-1vTPqglQkIVhni9NzkKeJUG_IYywFUFc1vNBK0j_TSDoK1S_WBkkgSlrRQ-xagjN44dVeAI8IU7krrLt/pubhtml}{full} dataset.

We see that \emph{GPT3.5} outperforms all other models on \emph{Human}, \emph{Wiki}, and \emph{Synthetic}, while the other LLM, \emph{FlanT5}, comes in second in the same datasets. The close performance between the models on \emph{Synthetic} implies that \emph{FlanT5} is a good approximation of \emph{GPT3.5} for this task, on boundaries generated by \emph{GPT3.5}. The biggest performance gap between the two is on \emph{Human}. Although this dataset is small, it represents a meaningful distribution shift and the smaller fine-tuned model is unable to generalize as well as the larger ChatGPT. This conclusion was also supported by manual inspection of segments.

Interestingly, we see that both BERT-based models are superior in the supposedly easier task posed by \emph{Conc-Wiki}. We theorise that these models are better at finding clear boundaries between different domains but struggle with more nuanced segment boundaries found in news articles, for example. By contrast, the LLMs are better at finding more nuanced segments that a human might have produced but are not significantly better at finding more clear-cut boundaries.

We also looked at precision and recall to better characterise the behavior of each model. \emph{GPT3.5} has the highest precision and recall on \emph{Human} and the highest recall on both \emph{Wiki} and \emph{Conc-Wiki}. This high recall is true in general for the LLMs, even on \emph{Conc-Wiki}, where the BERT models have higher overall boundary similarity. On the other hand, precision scores are generally closer, except on \emph{Conc-Wiki} where they are much better for the BERT models. This suggests that the BERT models are more hesitant in placing boundaries, but when they do, they are more likely to be correct. It should be noted that this behaviour in both the LLMs and BERT models is subject to the prompt and segment processing/validation procedures used.

\subsection{Qualitative Evaluation}

Through manual inspection of segmentations on \emph{Human}, we found that \emph{GPT3.5} found boundaries which seemed reasonable from a human perspective, especially for simpler documents like short news articles. \emph{FlanT5} model imitated this behavior but was less consistent. BERT segmenters would find reasonable segments, but after the manual gluing and splitting procedure, would often lead to off-by-1 errors and would miss some boundaries entirely.

For documents with more complex or nuanced text or with messy data like tables and artefacts, \emph{GPT3.5} would sometimes get stuck and return indices with a regular pattern that extends beyond the number of sentences in the input. For example, '$[1,15,22, \ldots, 76, 79, 82, 85, \ldots 118, 121, \ldots]$'. Better prompt engineering, a more rigorous data-processing procedure or the use of newer models might help. Our current approach was resource constrained but still required the ability to pass noisy documents to the model. A more thorough investigation of the logits computed by the model is required to understand how and when this occurs, and how to mitigate it.

% 4. Conclusion
\section{Conclusion}\label{Conclusion}
% 4. Conclusion

Our work compares generative LLMs with methods which use BERT embeddings and cosine similarity for topic segmentations. We propose a new prompting method that is token-efficient and guarantees that outputs are unedited. We support the use of boundary similarity for evaluation. Results indicate that LLMs can be more effective segmenters where more nuanced segmentations are required. However, when the input is noisy or the segment boundaries are clear, BERT-based methods may be more reliable. Future work should involve a thorough comparison of different prompting methods and address highlighted issues with LLM outputs using our method. Lastly, larger human-annotated datasets should be constructed rather than relying on headings or concatenated paragraphs.
% 5. Appendices
\section*{Ethics, Risks and AI Assistants}

There were two sources of data in this work. Wikipedia data is available under under the Creative
Commons Attribution-ShareAlike 4.0 International License (CC BY-SA). The other data sources are proprietary and cannot be made public. Code is also proprietary and cannot be made public.

All models used were accessed either by (paid) public API (\emph{GPT3.5}) or are open source models (\emph{FlanT5}, \emph{BERT}, \emph{BERTGraph}). The models were used in accordance with the terms of service of the respective providers. 

Potential risks of this work include the contribution to the desire for ever-more-powerful large language models, whose training and deployment can have negative consequences on the environment.

Github Copilot was used in writing code for this paper, but all code was reviewed and edited by the authors. The authors are responsible for the content of the paper and the code used in the experiments.
\section*{Limitations}

This work is comparable to a part of the work contained in~\cite{XingThesis}. However, their work is presented in the form of a PhD thesis which was made public in 2024. The experiments in this paper were conducted prior to this work being made public. This is part of the reason why none of our experiments directly compare our method to their prompting methods or loss-based approaches. The other reason is that we quickly reached the limit of our financial budget in the existing experiments. Further, our primary goal was to compare to the use of LLMs with the previously existing approach implemented at Adarga (who were funding this work). Finally, the code and datasets are no longer accessible to the authors due to their proprietary nature so we cannot rerun experiments on the same data with new prompting methods, and some details of experiments are no longer available to us. This is also why we unfortunately cannot release the data, code or models. Nonetheless, we hope that future work can directly compare our prompting method with those proposed in~\cite{XingThesis} or loss-based approaches on larger, publicly available datasets.

Results in this report are also subject to the subjective definition of a `segment' as implied by manual segmentation, wikipedia heading placement or examples in the few-shot prompt. The conclusions in this paper may be subject to this implicit definition of a segment, but we are optimistic that methods and results presented here are equally valid with more flexible definitions of segments and across different.

This work is also limited by the size and language of datasets used. The numbers reported are from a subset of \emph{Wiki} (only 150) due to the computational and financial resources available to us that were required to train and test the models. However, we tested on the full datasets with models for which it was computationally and monetarily feasible and found that the subset was a large enough sample to provide representative results. Full results have been made public\footnote{\href{https://docs.google.com/spreadsheets/d/e/2PACX-1vTPqglQkIVhni9NzkKeJUG_IYywFUFc1vNBK0j_TSDoK1S_WBkkgSlrRQ-xagjN44dVeAI8IU7krrLt/pubhtml}{link to full results}}. Finally, all datasets are in English and some of our ideas may not generalise to other languages. We hope that future work can be conducted on larger datasets incorporating more languages and domains. 
% \section*{Acknowledgments}

% This work was conducted during an internship at \href{https://adarga.ai/}{Adarga} in the summer of 2023. We would like to thank Adarga for the support and resources provided during the internship which allowed this project to take place. The two interns (Pierre and Maya) were supervised by Patrick. Pierre and Maya would like thank Patrick and other members of the data science and engineering teams at Adarga for their guidance and support throughout the project. We would also like to thank Hisham Alyahya for helpful feedback on the paper.

% Bibliography
\bibliography{custom}

\appendix
\section{Prompting Strategy}\label{Prompting Strategy}

The prompting strategy used in this work is a simple schema that is designed to be general and applicable to any LLM. The schema is as follows:

\begin{enumerate}
    \item The LLM is prompted with the input text, with integers in square brackets delimiting the sentence boundaries, few-shot examples of the task, a short instruction and a system prompt.
    \item Segments are validated. This means they must not be too long nor too short and that they do not contain too many punctuation marks as a proportion of the segment length.
    \item Segments that are too long are recursively split into smaller segments through similar prompting strategy. This prompt asks the LLM to return a single segment boundary index.
    \item This process is repeated until all segments are short enough.
    \item Segments that are too short are merged with a neighbouring segment based on the semantic similarity to neighbouring sentences. This part could also be done via prompting but we found this unnecessary.
\end{enumerate}

An example prompt is shown below.

\textbf{System:}

You are an expert linguist and a master of nuance in the meaning of written text. You are aware of when topics change in the flow of text and the meaning that words carry. You obey instructions. You think carefully before producing responses. You do not hallucinate. You are not a chatbot. You are not a summariser.

\textbf{Prompt:} 

You are given a document with sentence boundaries marked by square brackets. Your task is to segment the document into coherent parts. Return a list of indices corresponding to the segment boundaries of the document. This list should ONLY be a list of integers, for example '1, 3, 5'. Some examples are shown below.

Text: 

It was a sunny day in the park. [1] The birds were singing. [2] The children were playing. [3] The adults were chatting. [4] The dogs were barking. [5] The sun was shining. [6] The day was perfect. [7] However, then the rain came. [8] The children ran for cover. [9] The adults laughed. [10] The dogs howled. [11] The sun disappeared. [12] The day was ruined. [13] Fortunately, the next day was sunny again. [14] But it was actually too hot! [15] The children were sweating. [16] The adults were fanning themselves.

Segments: 

7, 13

\ldots \emph{[more examples]} \ldots

Text: 

The cat sat on the mat. [1] The dog sat on the floor. [2] The cat was black. [3] The dog was brown. [4] The cat was fluffy. [5] The dog was short-haired. [6] The cat was purring. [7] The dog was wagging its tail. [8] The cat was happy. [9] The dog was happy. [10] Then the cat went to London. [11] The dog went to Paris. [12] The cat saw the sights. [13] The dog saw the sights. [14] The cat ate fish and chips. [15] The dog ate croissants. [16] The cat drank tea. [17] The dog drank coffee. [18] The cat was happy. [19]

Segments:

\textbf{End Prompt}

We use a similar prompt for the recursive prompting mechanism with the same system prompt. For example:

\textbf{Recursive Prompt:}

You are given a document with sentence boundaries marked by square brackets. Your task is to choose one segment boundary to split the document into two coherent parts. Return a single integer corresponding to the index of the segment boundary. This integer should be between 1 and the number of sentences in the document. Some examples are shown below.

Text:

The cat sat on the mat. [1] The cat was black. [2] The cat was fluffy. [3] The cat was purring. [4] The cat was happy. [5] On the other hand, the dog sat on the floor. [6] The dog was brown. [7] The dog was short-haired. [8] The dog was wagging its tail. [9] The dog was happy. [10]

Segment:

5

\ldots \emph{[more examples]} \ldots

Text: 

Jack and Jill went up the hill. [1] Jack fell down and broke his crown. [2] Jill came tumbling after. [3] This is a well known nursery rhyme that has been passed down through the generations. [4] It is a classic. [5] It is a favourite of many. [6] It is a favourite of mine. [7] It is a favourite of yours. [8] It is a favourite of everyone.

Segment:

3

\textbf{End Prompt}

These examples are not illustrative of the length or style of segmentations in our dataset, they merely serve to exemplify the prompting schema. The actual prompts used in the experiments were much longer and more complex, and included more examples which were more realistic. The system prompt was also more detailed and included more examples of what the model should not do, such as not repeating the same segment boundary multiple times, not exceeding the length of the input sentences and not getting stuck in a pattern of regular segment boundaries.

% This document has been adapted
% by Steven Bethard, Ryan Cotterell and Rui Yan
% from the instructions for earlier ACL and NAACL proceedings, including those for
% ACL 2019 by Douwe Kiela and Ivan Vuli\'{c},
% NAACL 2019 by Stephanie Lukin and Alla Roskovskaya,
% ACL 2018 by Shay Cohen, Kevin Gimpel, and Wei Lu,
% NAACL 2018 by Margaret Mitchell and Stephanie Lukin,
% Bib\TeX{} suggestions for (NA)ACL 2017/2018 from Jason Eisner,
% ACL 2017 by Dan Gildea and Min-Yen Kan,
% NAACL 2017 by Margaret Mitchell,
% ACL 2012 by Maggie Li and Michael White,
% ACL 2010 by Jing-Shin Chang and Philipp Koehn,
% ACL 2008 by Johanna D. Moore, Simone Teufel, James Allan, and Sadaoki Furui,
% ACL 2005 by Hwee Tou Ng and Kemal Oflazer,
% ACL 2002 by Eugene Charniak and Dekang Lin,
% and earlier ACL and EACL formats written by several people, including
% John Chen, Henry S. Thompson and Donald Walker.
% Additional elements were taken from the formatting instructions of the \emph{International Joint Conference on Artificial Intelligence} and the \emph{Conference on Computer Vision and Pattern Recognition}.

\end{document}